\documentclass[conference]{IEEEtran}
\usepackage{cite}
\usepackage{amsmath,amssymb,amsfonts}
\usepackage{mathtools}
\usepackage{commath}
\usepackage{algorithmic}
\usepackage{bm}
\usepackage{graphicx}
\usepackage{textcomp}
\usepackage{xcolor}
\usepackage{color}
\usepackage{soul}
\usepackage{siunitx}
\usepackage{caption}
\usepackage{subcaption}
\def\BibTeX{{\rm B\kern-.05em{\sc i\kern-.025em b}\kern-.08em
    T\kern-.1667em\lower.7ex\hbox{E}\kern-.125emX}}
\DeclareMathOperator*{\argmin}{argmin}

\begin{document}

\title{EOS: a Parallel, Self-Adaptive, Multi-Population Evolutionary Algorithm for Constrained Global Optimization}

\author{\IEEEauthorblockN{Lorenzo Federici}
\IEEEauthorblockA{\textit{Department of Mechanical}  \\ \textit{and Aerospace Engineering} \\
\textit{Sapienza University of Rome}\\
Rome, Italy \\
lorenzo.federici@uniroma1.it}
\and
\IEEEauthorblockN{Boris Benedikter}
\IEEEauthorblockA{\textit{Department of Mechanical}  \\ \textit{and Aerospace Engineering} \\
\textit{Sapienza University of Rome}\\
Rome, Italy \\
boris.benedikter@uniroma1.it}\\   
 \and
\IEEEauthorblockN{Alessandro Zavoli}
\IEEEauthorblockA{\textit{Department of Mechanical}  \\ \textit{and Aerospace Engineering} \\
\textit{Sapienza University of Rome}\\
Rome, Italy \\
alessandro.zavoli@uniroma1.it}
}

\maketitle

\begin{abstract}
This paper presents the main characteristics of the
evolutionary optimization code named \textit{EOS}, Evolutionary Optimization at Sapienza, and its successful application to challenging, real-world space trajectory optimization problems.
\textit{EOS} is a global optimization algorithm for constrained and unconstrained problems of real-valued variables. It implements a number of improvements to the well-known Differential Evolution (DE) algorithm, namely, a self-adaptation of the control parameters, an epidemic mechanism, a clustering technique, an $\varepsilon$-constrained method to deal with nonlinear constraints, and a synchronous island-model to handle multiple populations in parallel.
The results reported prove that \textit{EOS} is capable of achieving increased performance compared to state-of-the-art single-population self-adaptive DE algorithms when applied to high-dimensional or highly-constrained space trajectory optimization problems.
\end{abstract}


\section{Introduction}
\textit{Evolutionary Optimization at Sapienza}, or \textit{EOS}, is an evolutionary optimization algorithm for continuous-variable problems developed
at the Department of Mechanical and Aerospace Engineering of Sapienza University of Rome.
Its origin dates back to 2015, when a first version of the algorithm was coded with the aim of tackling
the ``nearly-impossibile'' interplanetary trajectory optimization problems proposed in the Global Trajectory Optimization Competitions \cite{casalino2016gtoc8}.
Since then, \textit{EOS} has been continuously updated and improved, and applied with success to a broad range of unconstrained and constrained space trajectory optimization problems, as multiple gravity-assist trajectories \cite{federici2018preliminary}, 
rocket ascent trajectories \cite{Federici2019733, benedikter2019convex},
and multi-rendezvous missions \cite{federici2019impulsive, federici2019time}.

\textit{EOS} implements a multi-population, self-adaptive, $\varepsilon$-constrained Differential Evolution (DE) algorithm, with a synchronous island-model for parallel computation.
DE is a well-known population-based evolutionary algorithm, devised by R.~Storn and K.~Price in 1997 \cite{storn1997differential} to find the global optimum of nonlinear, non-differentiable functions of real-valued variables. 
Despite its simplicity, DE exhibits much better performance in comparison with several other metaheuristic algorithms on a wide range of benchmark and real-world optimization problems, defined over a continuous parameter space \cite{vesterstrom2004comparative}.
As other Evolutionary Algorithms (EAs) and Genetic Algorithms (GAs), DE exploits the crossover, mutation, and selection operators to generate new candidate solutions, or individuals, and to decide on their survival in successive generations.
Unlike traditional EAs and GAs, each mutated solution is generated as a scaled difference of a number of distinct individuals of the current population. This self-referential mutation has the desirable property to automatically adapt the different variables of the problem to their natural scale in the solution landscape, boosting the search potential of the algorithm \cite{das2016recent}.
All these evidences contributed to selecting DE as the optimization core of \textit{EOS}.

The standard DE algorithm is neither a globally nor a locally convergent algorithm, that is, there exist problem instances for which DE is not theoretically able to identify locally optimal solutions \cite{locatelli2015non}.
Nevertheless, DE has proven to be capable of attaining high quality results in many practical optimization problems. 
However, its performance typically drop in complex optimization environments, characterized by nonlinear constraints, numerous isolated local minima, and a high-dimensional search space, all typical characteristics of space trajectory optimization problems, which are the kind of problems \textit{EOS} has been designed to deal with.
For this reason, many researchers focused on modifying the standard DE algorithm in order to improve its effectiveness when applied to hard, constrained, global optimization problems. Several DE variants have thus been proposed, most of which are collected in \cite{das2011differential, das2016recent}. 
Four weaknesses of the classical DE algorithm have been particularly targeted by researchers: (i) (nonlinear) constraint handling \cite{coello1999survey}, (ii) the tuning of the few DE control parameters that drive the evolution process \cite{eiben1999parameter}, (iii) the lack in diversity between individuals in the population over generations \cite{lampinen2000stagnation}, and (iv) the need to find a balance between a wide exploration of the solution space and a quick refinement (or exploitation) of the previously obtained solutions \cite{eiben1998evolutionary}.

All these aspects have been addressed in \textit{EOS} by combining some of the most successful ideas found in the literature.
Specifically, five major add-ons have been implemented in \textit{EOS} to improve the performance of the standard DE algorithm: (i) a $\varepsilon$-constrained method, to deal with (possibly nonlinear) constraints, (ii) a self-adaptation of the control parameters, (iii) an epidemic mechanism, to maintain diversity within the population during the evolution, (iv) a pruning of the worst sections of the solution space, to speed up the convergence process, and (v) a synchronous, multi-mutation, island-model, to achieve a proper balance between exploration and exploitation of the search space. Moreover, the code has been made parallel by using a hybrid MPI-OpenMP programming, to obtain reasonable computation times even in presence of high dimensional problems and/or computationally expensive cost functions. 

The paper is organized as follows. After the basic notation of optimization problems is introduced in Sec.~\ref{sec:problem}, the standard DE algorithm and its operators are briefly described in Sec.~\ref{sec:DE}. The following sections give a detailed description of the improvements made in \textit{EOS} to DE, i.e., the self-adaptive scheme (Sec.~\ref{sec:selfadapt}), the epidemic mechanism (Sec~\ref{sec:epidemic}), the pruning technique (Sec~\ref{sec:pruning}), constraint handling (Sec.~\ref{sec:eps-con}) and the parallel island-model (Sec.~\ref{sec:islands}). 
Finally, successful applications of \textit{EOS} to real-world problems borrowed from the space sector, that is, the optimization of a Multiple Gravity-Assist (MGA) trajectory, of the ascent trajectory of a multi-stage launcher and of an Active Debris Removal (ADR) trajectory, are reported in Sec.~\ref{sec:test}.

\section{Optimization Problem}
\label{sec:problem}
Given a problem described by $D$ real-valued variables:
\begin{equation}
    \mathbf{x} = \left[x^{(1)}, \dots , x^{(D)}\right]
\end{equation}
and a cost function $f(\mathbf{x})$, the aim of an optimization process is to find the vector 
$\mathbf{x}^\ast$ that minimizes $f(\mathbf{x})$:
\begin{equation}
    \mathbf{x}^\ast = \argmin_{\mathbf{x}\in\Omega} f(\mathbf{x})
    \label{eq:optProb}
\end{equation}
where $\Omega$ represents the solution space.
In \textit{unconstrained optimization problems}, $\Omega$ is a $D$-dimensional hyperrectangle, defined as the Cartesian product of the bounding intervals of the design variables $\mathbf{x}_L = \left[x_L^{(1)}, \dots , x_L^{(D)}\right]$ and $\mathbf{x}_U  = \left[x_U^{(1)}, \dots , x_U^{(D)}\right]$:
\begin{equation}
   \Omega = \Omega_b = \left \{ \mathbf{x} \in \mathbb{R}^D : \mathbf{x}_L \leq  \mathbf{x} \leq \mathbf{x}_U \right\} \subset \mathbb{R}^D
   \label{eq:constr_1}
\end{equation}

In the case of \textit{constrained optimization problems}, the solution space $\Omega$ is further reduced by the presence of $K$ inequality constraints $\bm{\Psi}(\mathbf{x}) = \left[ \Psi^{(1)}(\mathbf{x}), \dots, \Psi^{(K)}(\mathbf{x}) \right]$, which in general are nonlinear functions of the design variables:
\begin{equation}
   \Omega = \left \{ \mathbf{x} \in \Omega_b :  \bm{\Psi}(\mathbf{x}) \leq \bm{0}  \right \} \subset \mathbb{R}^D
   \label{eq:constr_2}
\end{equation}
%
The possible presence of equality constraints is already included in expression~\eqref{eq:constr_2}; indeed, any equality constraint of the form: $\Phi^{(k)}(\mathbf{x}) = C$ can be rewritten as an inequality constraint if an arbitrarily small tolerance $\delta$ is introduced:
\begin{equation}
 \Psi^{(k)}(\mathbf{x}) \coloneqq |\Phi^{(k)}(\mathbf{x}) - C| - \delta \leq 0\\
 \label{eq:eqcon}
\end{equation}

\section{Standard Differential Evolution}
\label{sec:DE}
A brief description of the standard DE algorithm 
is here provided. 
Let us consider the unconstrained minimization problem defined by Eqs.~\eqref{eq:optProb} and \eqref{eq:constr_1}. 
An initial collection (or \textit{population}) $pop^0$ of $N_p$ candidate solution vectors (or \textit{individuals}) is generated by randomly sampling  the solutions, as evenly as possible, in the solution space:
\begin{equation}
    pop^0 = \{ \mathbf{x}_i \in \Omega \}_{i=1,\dots,N_p}
\end{equation}
where:
\begin{equation}
    x^{(j)}_i = x^{(j)}_L + p^{(j)}_i \left(x^{(j)}_U - x^{(j)}_L \right)
    \label{eq:DEinit}   
\end{equation}
for $j = 1,\dots,D$, where $p^{(j)}_i$ indicates a random number with uniform distribution in $[0,\, 1]$.
The value of the cost function (or \textit{fitness}) $f(\mathbf{x}_i)$ is evaluated for each individual (or \textit{agent}) $\mathbf{x}_i$ composing the initial population.

At any iteration $G$ (or \textit{generation}) of the algorithm, 
a new population $pop^{G+1}$ is created by applying to each vector $\mathbf{x}_i \in pop^G$ a sequence of three operations, named mutation, crossover, and selection, defined as follows.

\subsection{Mutation}
\label{sec:Mut}
During mutation, a mutated or \textit{donor} vector $\mathbf{v}_i$ is created as a linear combination of a few population members. Several mutation rules were proposed in the original paper by Storn and Price \cite{storn1997differential} in order to attain either a wider exploration of the search space or a faster convergence to the optimum (i.e., exploitation). Since then, many other rules have been devised, with the same purpose, by a number of researchers \cite{ali2011differential}.
In the current version of \textit{EOS}, four strategies, among those available in the literature, are adopted:
\begin{equation}
\begin{array}{ll}
    1) & \mathbf{v}_i = \mathbf{x}_{r_1} + F\,(\mathbf{x}_{r_2} - \mathbf{x}_{r_3}) \\ 
    2) & \mathbf{v}_i = \mathbf{x}_{best} + F\,(\mathbf{x}_{r_1} - \mathbf{x}_{r_2}) \\
    3) & \mathbf{v}_i = \mathbf{x}_{i} + F\,(\mathbf{x}_{r_3} - \mathbf{x}_{i}) + F\,(\mathbf{x}_{r_1} - \mathbf{x}_{r_2})\\
    4) & \mathbf{v}_i = \mathbf{x}_{best} + F\,(\mathbf{x}_{r_1} - \mathbf{x}_{r_2}) + F\,(\mathbf{x}_{r_3} - \mathbf{x}_{r_4})
\end{array}
\label{eq:DEmut}
\end{equation}
where $F \in \mathbb{R}$ is a parameter driving the mutation (\textit{scale factor}), $\mathbf{x}_{best}$ is the best individual in the current population, and $r_1, \dots, r_4$ represent randomly chosen, non-repeated, indexes belonging to $[1, N_p] \setminus \{i\}$.
Each strategy has its own weaknesses and strengths: strategies based on the mutation of the best individual, such as strategies 2 and 4 (referred to as \textit{DE/best/1} and \textit{DE/best/2}, respectively)
typically show a faster rate of convergence toward an (often local) minimum, whereas strategies based on the perturbation of a randomly chosen or the current individual,
such as strategies 1 and 3 (referred to as \textit{DE/rand/1} and \textit{DE/current-to-rand/1}, respectively) explore to a greater extent (yet more slowly) the whole search space.

\subsection{Crossover}
\label{sec:Cross}
During crossover, a \textit{trial} vector $\mathbf{u}_i$ is obtained by mixing the components of the agent vector $\mathbf{x}_i$ and the donor vector $\mathbf{v}_i$. By relying on empirical comparisons \cite{mezura2006comparative}, the \textit{binomial} crossover was preferred to the \textit{exponential} one in \textit{EOS}, and implemented as:
\begin{equation}
\begin{array}{cc}
     {u}^{(j)}_i = & 
     \begin{cases}
     {v}^{(j)}_i & \mbox{if}\,\, p^{(j)}_i \leq C_r\,\, \mbox{or}\,\, j = j_r\\
     {x}^{(j)}_i & \mbox{otherwise}
     \end{cases} 
\end{array}
\label{eq:DEcross}
\end{equation}
for $j = 1,\dots,D$, where $p^{(j)}_i$ is a random number with uniform distribution in $[0, 1]$, $C_r \in [0, 1]$ is another control parameter of the algorithm (named \textit{crossover probability}) and $j_r$ is a random index chosen once per individual in the range $[1, D]$ that ensures that at least one element in  $\mathbf{u}_i$ is inherited from $\mathbf{v}_i$.
At this point, the bounding intervals are enforced on the design variables:
\begin{equation}
\begin{array}{cc}
     {u}^{(j)}_i = & 
     \begin{cases}
     {x}^{(j)}_L & \mbox{if}\,\,\, {u}^{(j)}_i < {x}^{(j)}_L\\
     {x}^{(j)}_U  & \mbox{if}\,\,\, {u}^{(j)}_i > {x}^{(j)}_U\\
     {u}^{(j)}_i & \mbox{otherwise}
     \end{cases} 
\end{array}
\label{eq:DEbound}
\end{equation}
for $j = 1,\dots,D$.

\subsection{Selection}
\label{sec:Sel}
Eventually, the agent and trial vectors are compared.
In the case of an unconstrained optimization problem, the one that is characterized by the best fitness value is retained and inserted in the new population $pop^{G+1}$:
\begin{equation}
\begin{array}{lc}
     \mathbf{x}_i^{G+1} = & 
     \begin{cases}
    \mathbf{u}_i^G & \mbox{if}\,\,\, f(\mathbf{u}_i^G ) \leq f(\mathbf{x}_i^G )\\
    \mathbf{x}_i^{G} & \mbox{otherwise}
     \end{cases} 
\end{array}
\label{eq:DEsel}
\end{equation}
The corresponding selection step in the case of constrained optimization is discussed in Sec.~\ref{sec:eps-con}.

\subsection{Termination Criteria}
\label{sec:term}
This three-step process is repeated iteratively, creating at each generation a new population that replaces the previous one. 
Several termination criteria can be adopted, either alone or in conjunction, to stop the optimization procedure. The most common are:
\begin{itemize}
    \item maximum number of fitness evaluations ($FES$), $N_F$;
    \item maximum number of generations, $N_G$;
    \item maximum number of consecutive generations without any improvement of the best solution found, $N_{G, best}$.
\end{itemize} 

In typical \textit{EOS} applications, the termination criterion is
mainly based on the maximum number of generations $N_G$.
This parameter strongly depends
on the complexity of the analyzed problem, and it is generally selected in such a way that the outcomes
of independent runs of the algorithm bring, in almost all cases, to similar results.
So, a few preliminary runs of the code are necessary to identify a suitable value for $N_G$.

\section{Self-Adaptation of Control Parameters}
\label{sec:selfadapt}
A common practical issue for many stochastic algorithms concerns the selection of suitable values for the control parameters, i.e., the numerical quantities that drive the different phases of the optimization process.
A fine tuning of the parameters is often mandatory in order to maximize the algorithm performance while solving a given problem. 

The basic version of DE is characterized by only three control parameters. Apart from the population size $N_p$, the performance of the DE algorithm depends on an appropriate selection of the scale factor $F$, which drives the mutation phase, and the crossover probability $C_r$, which drives the crossover phase.
However, as for other metaheuristics, the best tuning of the control parameters, in terms of effectiveness and robustness of the algorithm, is usually related to the structure of the problem at hand \cite{liu2002setting}.

In order to avoid a manual trial-and-error tuning of the DE control parameters prior to the solution of any problem, the $j$DE self-adaptive scheme proposed by Brest et al.\cite{brest2007performance} is implemented in \textit{EOS} for automatically adjusting the values of both $F$ and $C_r$ during the optimization, without introducing any significant computational burden.
In $j$DE, each individual owns its private copy of $F$ and $C_r$, which are randomly initialized within the intervals $[F_{min}, F_{max}]$ and $[C_{r,min}, C_{r,max}]$, and thus different from individual to individual.
Therefore, each individual evolves according to its own set of parameters. The hope is that good values of these control parameters would contribute to produce better individuals, which, being more prone to survive and produce offspring, will propagate their parameters into the population on following generations.

At the end of the current generation $G$, each individual $\mathbf{x}_i$ undergoes, with a probability $p_\tau$, a random mutation of its control parameters:
\begin{equation}
\begin{array}{l}
     F_i^{G + 1} =  
     \begin{cases}
        F_{min} + p_{i,1}\Delta F & \mbox{if}\,\, p_{i,2} \leq p_\tau\\
      F_i^{G} & \mbox{otherwise}
     \end{cases} \\
     \\
     C^{G + 1}_{r,i} =
     \begin{cases}
        C_{r,min} + p_{i,3}\Delta C_r & \mbox{if}\,\, p_{i,4} \leq p_\tau\\
      C^{G}_{r,i} & \mbox{otherwise}
     \end{cases} 
\end{array}
\label{eq:SelfAdapt}   
\end{equation}
where $p_{i,1}, \dots, p_{i,4}$ are random numbers sampled from a uniform distribution in $[0, 1]$, $\Delta F = F_{max} - F_{min}$ and $\Delta C_r = C_{r,max} - C_{r,min}$. The hyperparameter values suggested in \cite{brest2007performance} are used in \textit{EOS}: $F_{min} = 0.1$, $F_{max} = 1$, $C_{r,min} = 0$, $C_{r,max} = 1$ and $p_\tau = 0.1$.

The population size $N_p$, instead, is kept fixed across generations. Its value is predetermined by looking at the problem dimension; a reasonable value for $N_p$ generally lies between $5D$ and $10D$. {For high-dimensional problems, e.g., $D>100$}, $N_p$ is more often selected on the basis of the available computational budget.

\section{Epidemic Mechanism}
\label{sec:epidemic}
A partial restart mechanism, named ``epidemic'', is adopted in \textit{EOS} in order to promote diversity between individuals of the population over the generations. In fact, the use of (guided) restart procedures in DE has demonstrated to be effective in reducing the chances of stagnation of the algorithm in isolated local minima \cite{peng2009multi, vasile2011inflationary}.

For this purpose, 
at each generation the population diversity score $\bar{d}$ is evaluated as the average Euclidean distance between any pair of solutions in the current population:
\begin{equation}
\bar{d} = \frac{2}{N_p (N_p - 1)}\sum_{i = 1}^{N_p} {\sum_{\substack{j = 1\\ j \neq i}}^{i} d_{ij}}
\label{eq:diversityscore}
\end{equation}
where:
\begin{align}
    d_{ij} &= \sqrt{ \sum_{k = 1}^{D} { \left(\frac{x_i^{(k)} - x_j^{(k)}}{ x_U^{(k)} - x_L^{(k)} }  \right)^2 }} \label{eq:diversity} 
    %
\end{align}
Thus, $d_{ij}$ would represent the actual (Euclidean) distance between the individuals $\mathbf{x}_i$ and $\mathbf{x}_j$ if the solution space were a $D$-dimensional unitary hypercube. For this reason, $d_{ij} \in [0, \sqrt{D}]$.

\begin{figure} [htbp]
    \centering
    \includegraphics[width = 0.8\columnwidth]{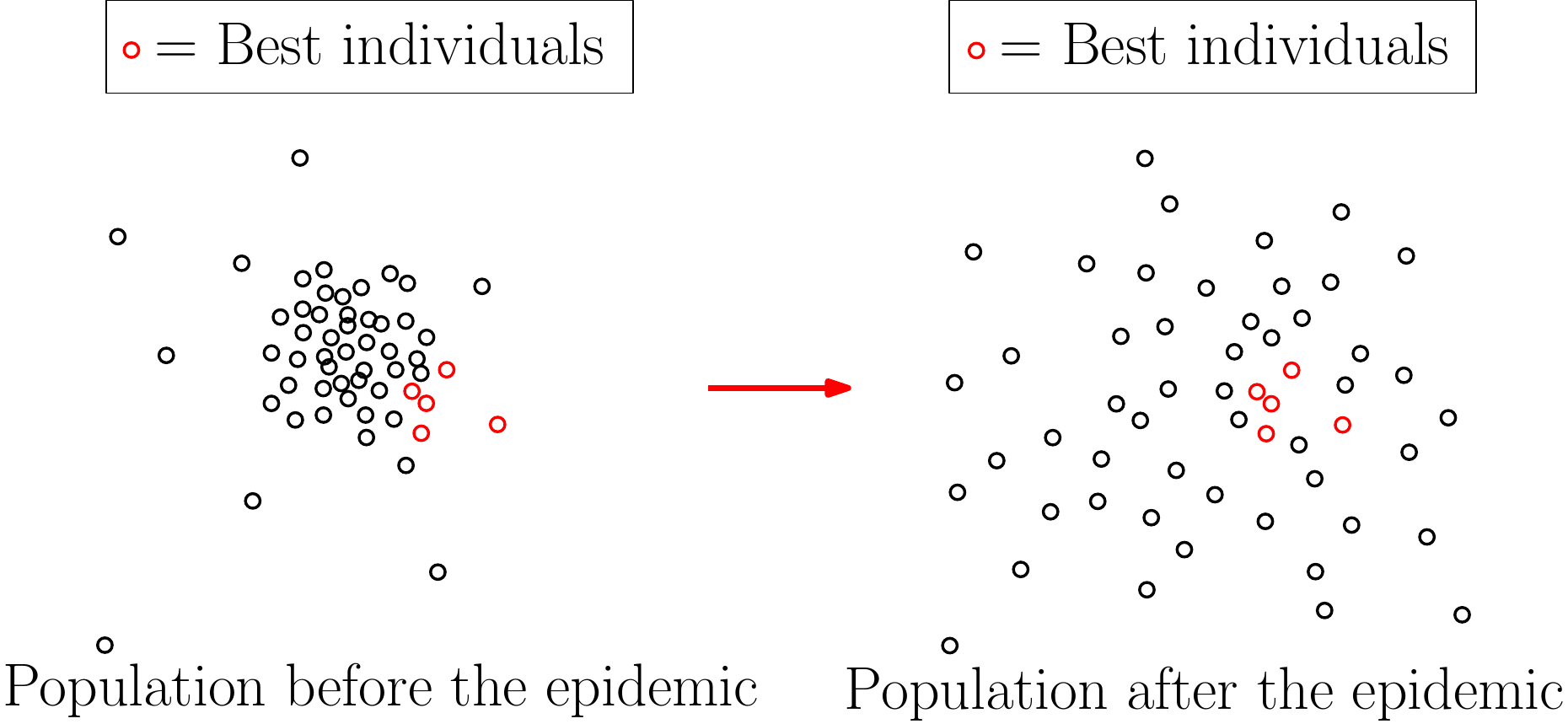}
    \caption{Effect of the epidemic mechanism on the population.}
    \label{fig:epidemic}
\end{figure}

When the diversity score $\bar{d}$ falls under a given small threshold $d_{tol}$, an epidemic unleashes on the population. 
The best $\rho_{elite} N_p$ individuals are immune to the epidemic; instead, a large portion $\rho_{ill}$ of the remaining individuals, randomly selected, contracts the ``fatal disease'', that is, is randomly reinitialized in the whole search space.
This mechanism is illustrated in Fig.~\ref{fig:epidemic}.
The epidemic cannot happen twice within a number $N_{G, epid}$ of consecutive generations, so as not to compromise the search.
Reasonable values for the newly introduced hyperparameters can be chosen in the following ranges: $N_{G, epid} = 500 \div 2000$, $d_{tol} = 10^{-4} \div 10^{-2} $, $\rho_{elite} = 0.05 \div 0.25$, $\rho_{ill} = 0.75 \div 1$.

\begin{figure} [htbp]
    \centering
    \includegraphics[width = 0.8\columnwidth]{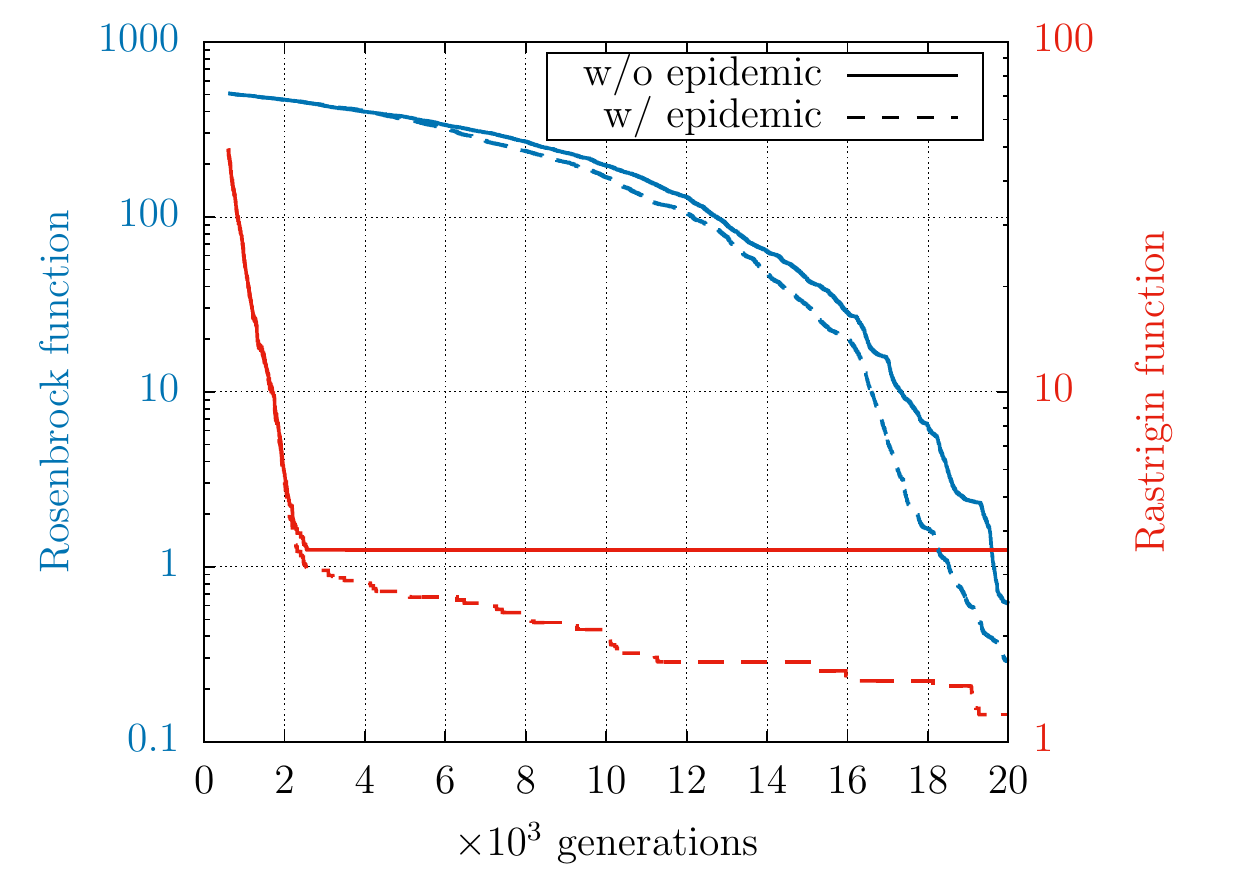}
    \caption{Evolution plots, averaged on 20 independent runs, obtained with and without the epidemic mechanism on two benchmark functions.}
    \label{fig:epidemic_benchmark}
\end{figure}

The epidemic mechanism has proven to be particularly effective when an exploitative mutation rule (e.g., strategy 2 or 4) is chosen for DE.
Figure~\ref{fig:epidemic_benchmark} shows the effectiveness of the mechanism on two standard benchmark functions, that is, Rosenbrock function, with $D = 100$ and bounding intervals $[-50,50]^D$, and Rastrigin function, with $D = 30$ and bounding intervals $[-5.12, 5.12]^D$.
The plot reports the change in fitness trend (averaged on 20 independent runs) obtained when the epidemic mechanism is added to a single-population self-adaptive DE algorithm, where a self-adaption rule analogous to those in Eq.~\ref{eq:SelfAdapt} was devised also to decide the mutation strategy among the four ones reported in Sec.~\ref{sec:Mut}.
The results were obtained with the following algorithm settings: $N_p = 64$, $N_G = 20000$, $N_{G, epid} = 1000$, $d_{tol} = 10^{-3}$, $\rho_{elite} = 0.1$, $\rho_{ill} = 1$.

\section{Space Pruning by Clustering}
\label{sec:pruning}
The term ``space pruning'' refers to those techniques that aim at reducing the solution space during the search,
focusing the optimization in a smaller area where, in accordance with some criteria, good solutions are expected to be found.

Different solution methods have been proposed in the past \cite{vasile2009hybrid, ghojogh2018tree} that combine branching, that is, the subdivision of the feasible region into smaller subdomains (with some convergence properties) and population-based stochastic algorithms, to search in the subdomains in order to evaluate them, showing higher average performance with respect to a number of stochastic and deterministic methods on the same problems.

The pruning method adopted in \textit{EOS} is based on the \textit{cluster pruning algorithm} developed by ESA's Advanced Concept Team \cite{izzo2010global} to solve Multiple Gravity-Assist (MGA) problems.
The key idea is that in problems like MGA and related space trajectory optimization problems (as MGA-1DSM or multi-rendezvous ones), which are the kind of problems addressed by \textit{EOS}, good solutions are often clustered in small regions instead of
being densely distributed over the whole search space.
So, during the optimization process, it is possible to progressively focus the search around the promising regions identified so far, increasing the quality of the solutions found, but accepting the risk that some possibly better solution may be potentially ruled out.

The \textit{pruning-by-clustering} method implemented in \textit{EOS} 
makes use of $N_r$ independent, separate, partial runs.
Each partial run corresponds to a run of the same DE algorithm with a (different) random initial population.

At each pre-assigned generation $N^i_{G,pr}$, the $i$-th pruning event occurs.
The $N_r$ runs are stopped, and the best solution found by each run is collected in the set
$pop^{best}=\left\{\mathbf{x}_1,\,\mathbf{x}_2,\,\ldots,\mathbf{x}_{N_r}\right\}$, which is sorted according to fitness, so that $f(\mathbf{x}_1) \le f(\mathbf{x}_2) \le \ldots {\le f(\mathbf{x}_{N_r})}$.
A new (smaller) search space $\Omega^i_{pr}$ is defined
as the convex hull containing the best $N_{p, pr}^i = \lfloor \rho_{pr}^i N_r \rfloor$ solutions in $pop^{best}$, up to some relaxation factor related to $\rho_{pr}^i$.
More precisely,
for each design variable $x^{(j)}$, with $j \in [1,D]$, the new bounding interval $\left[x_{L, new}^{(j)}, x_{U, new}^{(j)}\right]$ is defined as:
\begin{align}
    x_{L, new}^{(j)} &= \min_{k \in [1, N_{p,pr}^i]}{x_k^{(j)}} - 0.5 (1 - \rho_{pr}^i) \left (x_U^{(j)} - x_L^{(j)} \right ) \\
    x_{U, new}^{(j)} &= \max_{k \in [1, N_{p,pr}^i]}{x_k^{(j)}} + 0.5 (1 - \rho_{pr}^i) \left (x_U^{(j)} - x_L^{(j)} \right )
\end{align}
where
$\rho^i_{pr} = \rho_{pr}^0 - i \Delta \rho_{pr}$ is a tuning parameter that controls the extent of the search space region to be removed.
After the pruning event, the $N_r$ independent, partial optimization runs are restarted, randomly initializing the population over the pruned search space $\Omega^i_{pr}$.
Nevertheless, the $N^i_{p, pr}$ best individuals of $pop^{best}$ are copied 
into each new population, substituting the (randomly generated) worst individuals.

If $N_{pr}$ pruning events are enforced, starting from generation $N^0_{G,pr}$,
the subsequent pruning events are spaced from each other by $(N_{G} - N_{G, pr}^0)/N_{pr}$ generations.
By choosing $N^i_{p, pr} = \lfloor \rho^i_{pr} N_p \rfloor$, one 
makes the pruned search space iteratively smaller, as the convex hull is created on less points.

%

The hyperparameter values that showed the best performance in EOS are the following: $\rho_{pr}^0 = 0.3$, $\Delta \rho_{pr} = 0.1$, $N_{pr} = 3$ and $N_{G, pr}^0 = 0.4N_G$. 

\section{Constraint Handling}
\label{sec:eps-con}
In order to tackle  the constrained optimization problem defined by Eqs.~\eqref{eq:optProb} and \eqref{eq:constr_2} with DE, it is necessary to modify in some way the selection step \eqref{eq:DEsel} to take into account the presence of inequality constraints.

The simple, but promising, min-max approach proposed by Jimenez and Verdegay \cite{jimenez1999evolutionary} is implemented in \textit{EOS} to handle constraints.
The key idea is to adopt a lexicographic order in the selection process, in which the constraint violation precedes the
objective function:
\begin{enumerate}
    \item between two feasible individuals, select on the basis of the minimum value of the cost function;
    \item between a feasible and an infeasible individual, select the feasible one;
    \item between two infeasible individuals, select on the basis of the lowest maximum constraint violation:
    \begin{equation}
        \Psi_{max} = \max\limits_{j \in [1, K]} \Psi^{(j)}(\mathbf{x})
    \end{equation}
\end{enumerate}
However, as highlighted by Coello \cite{coello1999survey}, this simple approach pushes the search to focus first only on the constraint satisfaction; so, if the feasible solution space is disjoint, the search could get trapped in a part of the feasible region that, in general, could be quite far from the global minimum of the problem, and, from which, it is impossible to escape.

An efficient way to overcome such drawback has been proposed by Takahama and Sakai \cite{takahama2010constrained}, who applied an $\varepsilon$-constrained method to Differential Evolution (``$\varepsilon$DE'').
They suggested introducing a tolerance $\varepsilon$ for constraint violation $\Psi_{max}$, to be decreased along generations. The following rule has been adopted in \textit{EOS} for evaluating $\varepsilon$ at generation $G$:
\begin{equation}
\begin{array}{cc}
     \varepsilon^{G} = &  
\begin{cases}
 \varepsilon^{0} & \mbox{for}\,\,\, G \leq N^{0}\\
 \varepsilon^{0} \left[\frac{\varepsilon^{\infty}}{\varepsilon^{0}}\right]^{\frac{G - N^{0}}{N^{\infty} - N^{0}}} & \mbox{for}\,\,\, N^{0} < G < N^{\infty}\\
 \varepsilon^{\infty} & \mbox{for}\,\,\, G \geq N^{\infty}
\end{cases}
\end{array}
\label{eq:eps}    
\end{equation}
with $\varepsilon^{0}, \varepsilon^{\infty}$ the initial and final values of the tolerance. As for $N^{0}$ and $N^{\infty}$, which define the interval in which $\varepsilon$ decreases, the following values showed the best overall results: $N^0 = \frac{N_G}{6}$, $N^\infty = N_G$.

By using Eq.~\eqref{eq:eps}, moves toward infeasible solutions are allowed at the beginning of the search, when the tolerance $\varepsilon$ is maximum and the entire solution space must be explored in order to identify promising regions. As the generation number increases, such moves become forbidden, and the search concentrates in the feasible part of the identified region, which, thanks to the initial search space exploration, is more likely to be closer to the optimum of the problem.

\section{Island-Model}
\label{sec:islands}
During the search process, a proper balance between two opposite needs, namely ``Exploitation'' and ``Exploration'', is paramount to the performance of any EA
 \cite{eiben1998evolutionary}.
Here exploitation refers to the capability of the evolutionary algorithm to exploit the information already collected from the population to focus the search toward the goal; exploration, instead, refers to the ability to introduce new information about the solution space into the population.
In DE, each mutation strategy (see Eq.~\eqref{eq:DEmut}) may privilege, to a greater extent, one of the two tendencies over the other \cite{feoktistov2006differential}. 
%
It goes without saying that certain mutation strategies perform better on some optimization problems than others, but the opposite may be true on different problems. 
%
This leads to the idea of combining different strategies together within a single search process, in order to obtain a more robust and performing algorithm, capable of successfully tackling a wider range of problems \cite{epitropakis2008balancing}.
In the same fashion as described in Sec.~\ref{sec:selfadapt} for the DE control parameters,
a self-adaptation of the mutation strategy could be devised \cite{qin2008differential}; unfortunately, this approach suffers from the fact that greedy (i.e., exploitative) strategies tend to prevail over the others in just a few generations.

\textit{EOS}, instead,  adopts a synchronous island-model paradigm as a way to 
concurrently handle different mutation strategies within one optimization run.
An island-based EA relies on the definition 
of several sub-populations, or ``tribes'', each one evolving independently of the others, according to its own (preassigned) algorithm. Each tribe lives on a separate ``island'', and all the islands are arranged in a cluster, or ``archipelago'', of arbitrary topology. 
Information sharing among the islands occurs only in sporadic events called ``migrations'', during which the best individuals move from an island to the neighboring ones, in accordance with the predefined archipelago topology \cite{martin1997c6}.
Numerical tests of multi-population DE-based algorithms on a broad range of low-to-high dimensional optimization problems \cite{tasoulis2004parallel, di2019adaptive} showed an improvement in performance with respect to sequential versions of the algorithm.

\begin{figure} [htbp]
    \centering
    \includegraphics[width = 0.8\columnwidth]{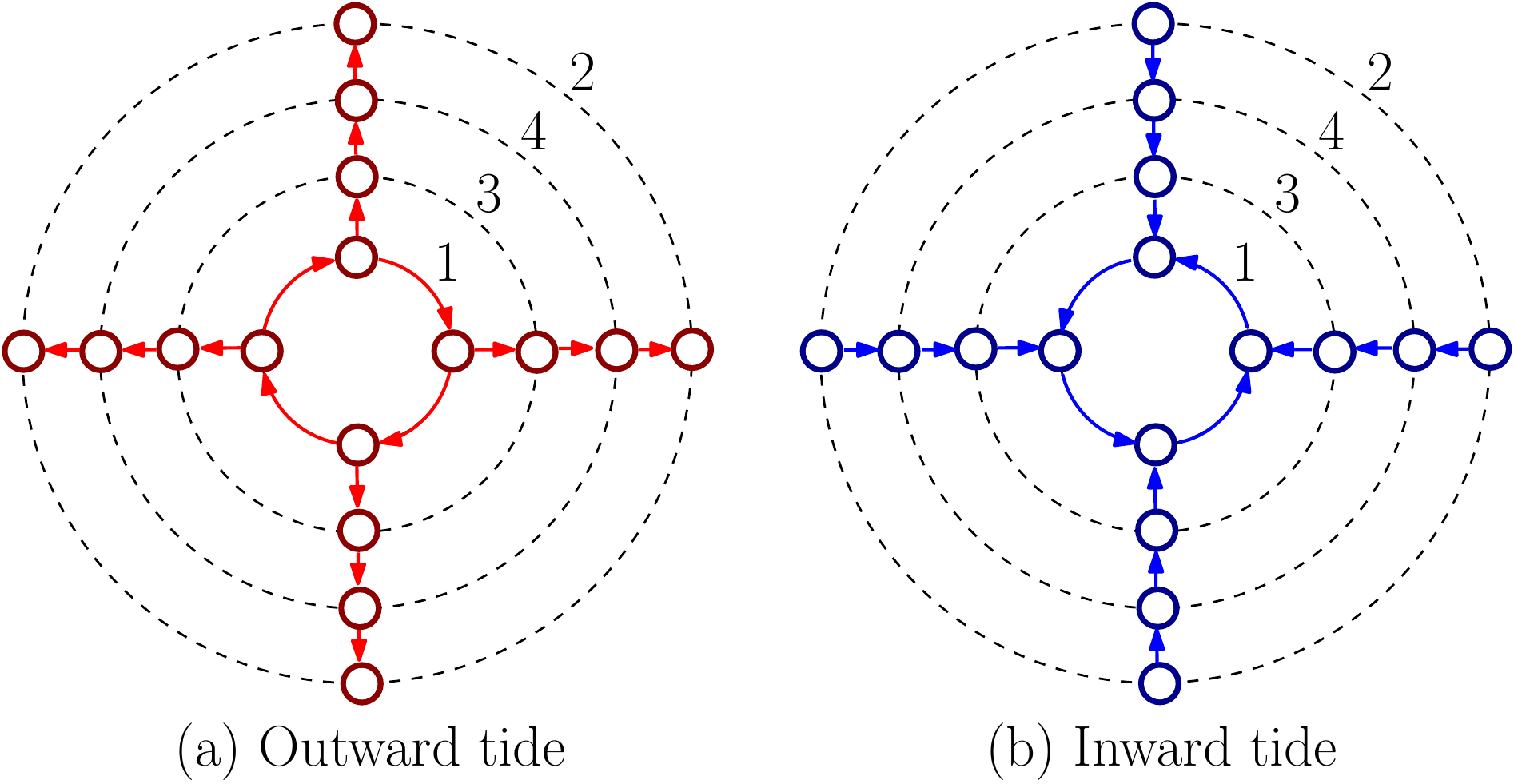}
    \caption{Migration tide: forward (a) and backward (b), for the 16-island case.}
    \label{fig:island}
\end{figure}

The island-model paradigm can be exploited to combine the convergence and search properties of different DE algorithms.
The idea is that by using heterogeneous mutation strategies among different islands it is possible to achieve a correct balance between the search space exploration and exploitation and perform better than the best of the strategies involved in that particular problem \cite{izzo2009parallel}.
As an example, in the radially-arranged 16-island archipelago in Fig.~\ref{fig:island}, inner rings favor exploration, featuring mutation strategies $1$ and $3$, while outer rings favor exploitation, featuring strategies $2$ and $4$. 

The island-model optimization process, with its alternating phases of extended, isolated computation and occasional communications, can be easily parallelized on a message-passing multi-processor environment.
The MPI message-passing standard \cite{clarke1994mpi} 
is exploited in \textit{EOS} for this purpose: each of the $N_i$ islands corresponds to a process, and is assigned to a different node/CPU of a cluster. The evolution phase proceeds in parallel, until communications between processes are performed during migrations.
More precisely, a synchronous migration of the best $N_b = \rho_{mig}N_p$ individuals occurs between the connected islands every $N_{mig}$ generations, with a probability $\phi_{mig}$. The $N_b$ best individuals are copied in the destination islands, where they replace the $N_b$ worst individuals. Typical values of these parameters are: $\rho_{mig} = 0.05 \div 0.1$, $N_{mig} = 100$, $\phi_{mig} = 0.5 \div 1$.
Communications are easily handled through the point-to-point send/receive functions of the MPI library. In addition, MPI allows for handy implementation of Cartesian or graph process topologies, which define the ``neighbors'' of any process to/from which it can send/receive information; by exploiting such capability, it is possible to arrange the archipelago in any topology in a straightforward manner. 
Besides classical topologies, such as the ring topology or the Von Neumann grid topology \cite{lynn2018population}, 
\textit{EOS} implements as a default a peculiar radial topology, 
where migration tides alternate their direction at each event, as presented in Fig.~\ref{fig:island}. 

This coarse-grained parallelization, in which each tribe is mapped to a CPU, can be combined with a fine-grained parallelism, in which each individual (or group of individuals) of each tribe is assigned to a different core of the CPU \cite{Alba2002Parallelism}.
The fine-grained parallelization of each sub-population is realized in \textit{EOS} through OpenMP, 
which is a set of compiler directives and library routines used to perform shared-memory parallelism, e.g., between the cores of a single, multi-core CPU; so, if $n_c$ cores are available for each CPU of the cluster, $N_p/n_c$ individuals are mapped to each core through an OpenMP directive. 


\section{Application to real-world space trajectory optimization problems}
\label{sec:test}

This section presents the main results achieved by \textit{EOS} when applied to challenging, real-world space trajectory optimization problems. 
More detailed information about how the three optimization problems that follow have been formulated (that is, their objective function, design variables, bounding intervals and constraints) are reported in the corresponding references \cite{federici2018preliminary, Federici2019733, federici2019time}.


\subsection{Europa Tomography Probe Mission Design}
\label{sec:ETP}

In 2015, a scientific and engineering team at Sapienza University of Rome, in collaboration with the Imperial College of London, carried out a feasibility study for a probe
that could be launched as a piggyback payload for the NASA Europa Clipper mission, to enhance its scientific return \cite{Notaro2016}.
An innovative mission concept was proposed, where a small Europa's orbiter, named Europa Tomography Probe (ETP), hosting just one magnetometer and a transponder required for the Inter-Satellite Link (ISL) with the mother spacecraft, is proved to be capable of providing crucial information on the moon's interior structure, such as depth and conductivity of the subsurface ocean. Also, ISL supports the reconstruction of the mother spacecraft orbit, significantly improving the accuracy of the topographic reconstruction of Europa’s surface \cite{di2019augmenting}.

The optimization of ETP capture trajectory was crucial for the validity of the overall proposal, as a tight requirement on the  available propellant mass was enforced by the need to stay within a total probe mass specified by NASA.
%
%
A mission strategy based on the $v_\infty$ leveraging concept \cite{sims1994analysis} and the use of resonant orbits to exploit multiple gravity-assists from Europa was thus proposed \cite{federici2018preliminary}.
Under the assumption of a patched-conic model, with  
radius of the sphere of influence of the secondary bodies and travel time inside these regions negligible, and an impulsive-thrust model, 
the problem can be posed as an unconstrained optimization problem, also known as Multiple Gravity-Assist with One Deep Space Maneuver (MGA-1DSM) \cite{vasile2006preliminary}, where the objective is to minimize the overall $\Delta V$, which is the cumulative velocity variation performed by means of the on-board propulsive system.

A velocity formulation \cite{ceriotti2010global} of the problem was exploited: the overall capture trajectory is made up of a series of body-to-body legs, each starting with a flyby and composed of two ballistic arcs, a \textit{propagation arc} and a \textit{Lambert arc}, joined by an impulsive maneuver. 
The $k$-th leg of the trajectory is parameterized by using four parameters: $\{ r_{\pi,k}, \beta_k, \Delta T_k, \eta_k\}$ which represent,
respectively, the flyby radius, the flyby plane orientation, the leg flight-time, and the fraction of the leg flight-time
at which DSM occurs. By properly selecting the bounds for variables $\Delta T_k$ and $\eta_k$ it is possible to enforce a given resonance of the probe with a Jovian moon on the \mbox{$k$-th} leg.
An initial Lambert arc, which moves the probe from the assigned initial
condition (apoapsis of a Jovian orbit in $4$:$1$ resonance with Europa) to the first encounter with Europa, completes the formulation.
This initial leg is described by three variables: the probe release epoch $t_0$, and the flight time $\Delta T_0$ and flight angle $\Delta \theta$ along the leg.

\begin{figure}[htbp]
    \centering
    \includegraphics[width = 0.9\columnwidth]{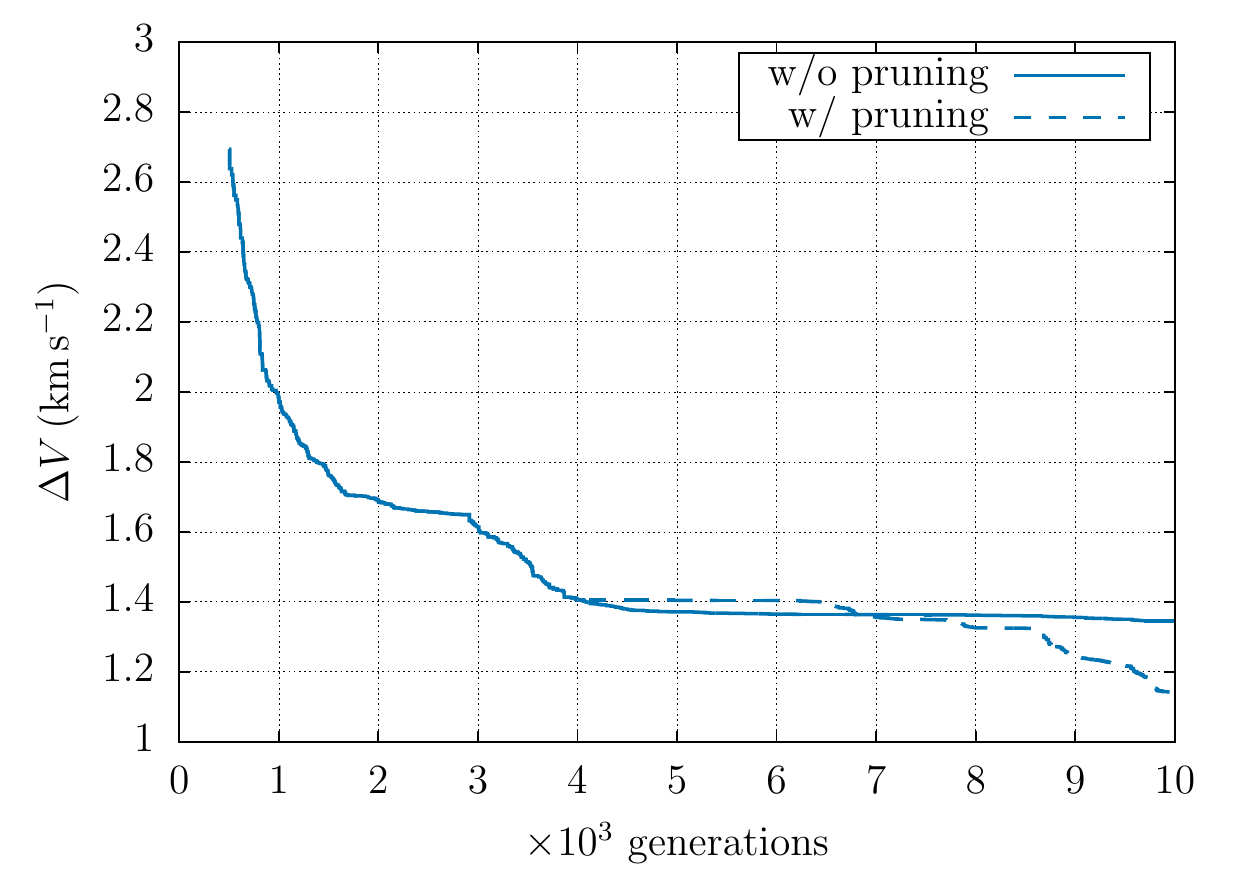}
    \caption{Best fitness trend obtained with and without the use of the pruning procedure. }
    \label{fig:pruning_MGA}
\end{figure}

MGA-1DSM problems are typically characterized by a huge number of local optima \cite{izzo2007search}; in this respect,
either the pruning-by-clustering method (Sec.~\ref{sec:pruning}) or the island-model (Sec.~\ref{sec:islands})
are paramount to the success of the optimization procedure through \textit{EOS}.
In particular, Fig.~\ref{fig:pruning_MGA} compares the results obtained on Mission C of Ref.~\cite{federici2018preliminary} by performing 50 runs of a mono-population, self-adaptive DE algorithm with and without the pruning-by-clustering procedure. The evolution plots reported are obtained by considering, at any generation number, the best fitness among those found by the separate runs.
The following algorithm settings were used: $N_i = 1$, $N_p = 64$, $N_G = 10000$, $N_r = 50$, $\rho_{pr}^0 = 0.3$, $\Delta \rho_{pr} = 0.1$, $N_{pr} = 3$ and $N_{G, pr}^0 = 0.4N_G$.

\begin{figure} [htbp]
    \centering
    \includegraphics[width=0.7\columnwidth]{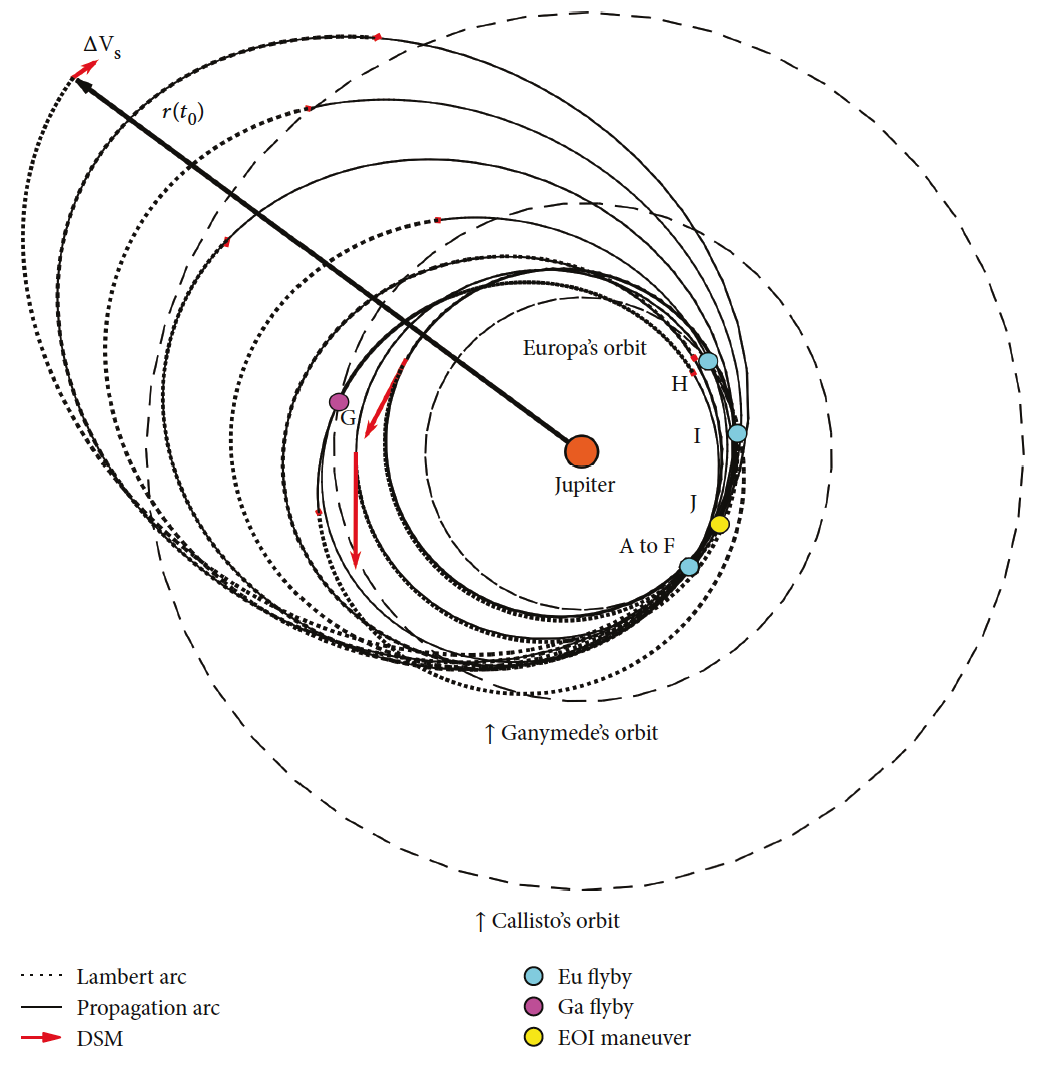}
    \caption{ETP capture trajectory.}
    \label{fig:ETP}
\end{figure}
%
The overall-best capture trajectory obtained for ETP is reported in Fig.~\ref{fig:ETP}, and allows saving approximately $\SI{1900}{\metre\per\second}$ of $\Delta V$ with respect to a direct insertion maneuver.
The trajectory exploits eight flybys of Europa and a flyby of Ganymede, for a total of 39 design variables.
The result was obtained with the following algorithm settings: $N_i = 8$, $N_p = 512$, $N_G = 10000$.

\subsection{Ascent Trajectory Optimization of VEGA Launch Vehicles}
\label{sec:VEGA}

Since 2017, \textit{EOS} is adopted as a solver for the optimization of the ascent trajectory of VEGA Launch Vehicle (LV) 
and its evolution VEGA-C 
in the framework of independent support and cross-check activity for ESA-ESRIN \cite{SoW2017}. 
By applying a control discretization approach, the problem of optimizing the ascent trajectory of a LV from the launch pad to a target orbit can be posed as a global constrained optimization problem.
This problem is highly sensitive to the value of the optimization variables and often the trajectory cannot be evaluated over the whole search space. Therefore, the use of a derivative-free optimization algorithm is mandatory.
With respect to the previous application, here the constraint handling plays a major role in determining the success or failure of the optimization process.
In particular, both the terminal constraints on the final LV orbital parameters and the path constraints limiting the dynamic pressure, the bending and axial loads experienced by the rocket, and the heat flux on the payload during the flight after the fairing jettisoning must be considered.
As a result, the admissible portion of the search space is quite small. 
%
%
Traditional approaches of constraint handling for EAs based on penalty functions \cite{richardson1989some} are not very effective in this case, because they tend to create, and get the search trapped into, a number of spurious sub-optimal solutions. Moreover, they are strongly reliant on a proper choice of the constraint weighting factors.
Barrier methods are not of any help too, as the feasible set in the search domain is quite limited and disconnected.

\begin{figure} [htbp]
    \centering
    \includegraphics[width = 0.9\columnwidth] {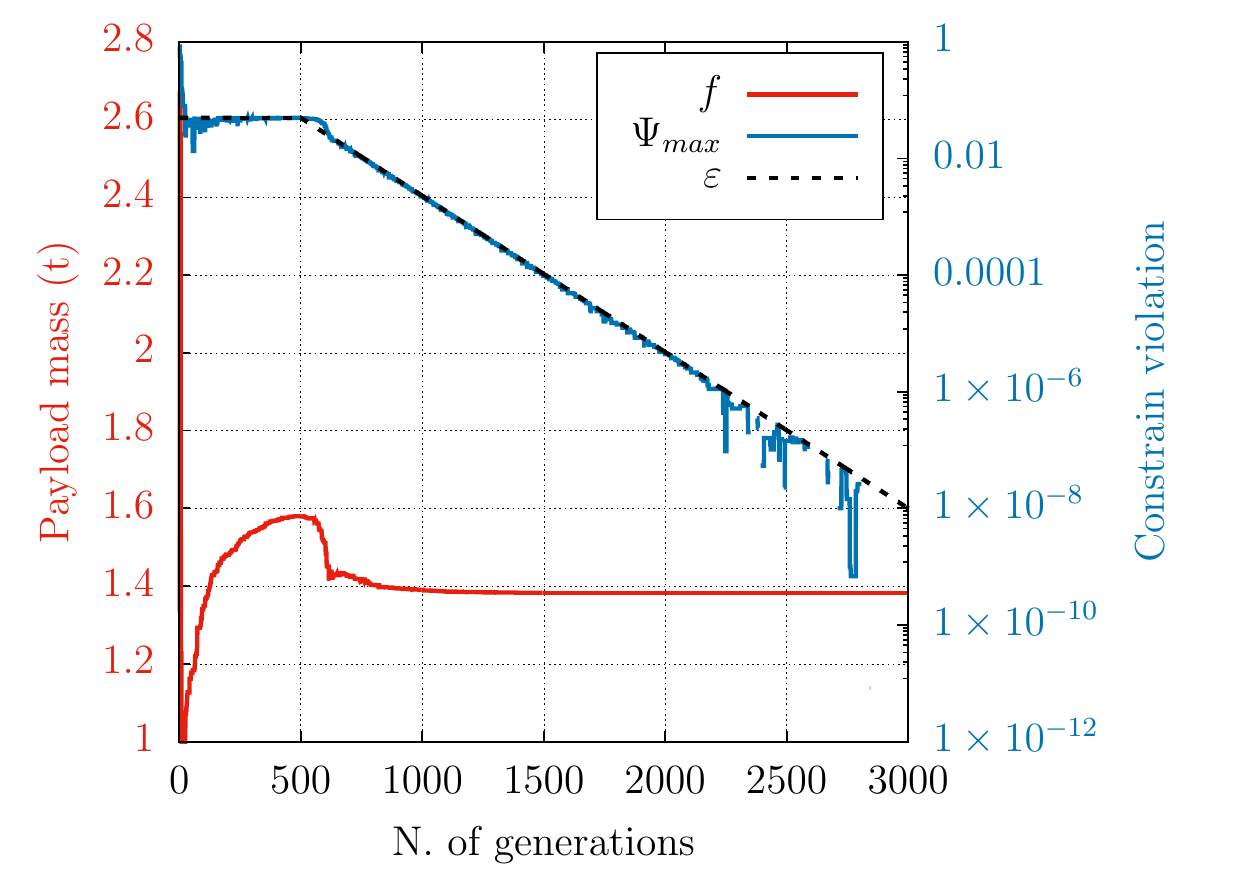}
    \caption{Typical trend of fitness $f$, constraint violation $\Psi_{max}$ and $\varepsilon$ level during a run of \textit{EOS} for an ascent trajectory optimization problem.}
    \label{fig:eps_vega}
\end{figure}

On the other hand, the $\varepsilon$-constrained method adopted in \textit{EOS} is really effective on this kind of problems, since starting from a value of the tolerance $\varepsilon$ sufficiently high, $\varepsilon$-feasible solutions (i.e., which meet the imposed constraints with a tolerance $\varepsilon$) can be obtained very soon during the search. Then, by decreasing in a slow and monotone way the parameter $\varepsilon$, the EA is able to maintain the solution feasible and, at the same time, attain better values of the fitness. As a result, at the end of the search, when $\varepsilon$ is close to zero, a feasible, good quality solution is returned by the algorithm. This process is shown in Fig.~\ref{fig:eps_vega} for VEGA LV, by using as objective function (to maximize) the payload mass $M_u$.

A typical run for a low/medium-fidelity ascent trajectory optimization with 22 variables and 9 nonlinear inequality constraints ($N_i = 1$, $N_p = 128$, $N_G = 3000$) requires about $5$ minutes
on a workstation with Intel Core i9-9900K @3.60 GHz,
using a single MPI process and up to 16 OpenMP threads.

An integrated optimization of the ascent trajectory and first stage Solid Rocket Motor (SRM) design of a VEGA-like multi-stage launch vehicle was also carried out through \textit{EOS}, where the objective was to determine the optimal internal pressure law during the first stage SRM operation, together with the optimal thrust direction
and other relevant flight parameters of the entire ascent trajectory, so as to maximize the payload injected into a
target orbit. Multiple design constraints involving the solid rocket motor or dependent on the actual flight trajectory were also enforced.
Here the problem was even more complex, because of the large number of constraints and its multi-disciplinarity (the design variables represent very distant physical quantities); nevertheless, \textit{EOS} was still able to attain competitive results  \cite{Federici2019733}.

In this case, a typical high-fidelity optimization ($N_i = 8$, $N_p = 136$, $N_G = 10000$), with 22 variables and 12 nonlinear inequality constraints, has a run time of about 30 minutes on KNL partition of  CINECA’s  supercomputer  Marconi  (3600  compute  nodes with 68-core Intel(R) Knights Landing @1.40GHz and 16 GBMCDRAM  +  96  GB  RAM  each,  connected  through  an  IntelOmniPath network @100Gb/s), by using 8 nodes and 68 CPU cores per node.

\subsection{Active Debris Removal Missions}
\label{sec:ADR}
An Active Debris Removal (ADR) mission can be seen as a peculiar instance of 
a multi-rendezvous (MRR) trajectory, 
where  an active (chaser) spacecraft is 
asked to visit
(that is, to perform a rendezvous with)
a certain number of targets (space debris), making the best use of the on-board propellant.

The optimization of a multi-target rendezvous trajectory can be posed as  
a mixed-integer NLP problem, involving the simultaneous optimization of both integer variables (defining the debris encounter sequence) and real-valued variables (describing the spacecraft trajectory from a debris to the next one and the encounter epochs).
A bi-level approach can be pursued for solving this NP-hard problem,
by isolating 
i) an outer-level, which concerns the definition of the encounter sequence and approximate encounter epochs, based on a rough, but fast, evaluation of each debris-to-debris cost (an \emph{heuristic}), and
%
ii) an inner-level, which deals with the optimization of each body-to-body transfer with full details, assuming that departure and arrival bodies are assigned; encounter epochs are also adjusted.


Given a proper heuristic to estimate the cost of each transfer leg (without solving the full optimization problem), the two levels can be solved sequentially: 
the outer-level combinatorial problem is isolated and solved first, for example
using either a Genetic Algorithm \cite{federici2019impulsive} or Simulated Annealing \cite{federici2019time};
its solution is then used as initial guess for the inner-level NLP problem, that is the actual trajectory optimization problem. 

If a Keplerian dynamical model is assumed for both the targets and the chaser, the inner-level problem results in an unconstrained problem of real-valued design variables with the total $\Delta V$ as cost function, which can be tackled by \textit{EOS}.
When the $N$ targets to remove are supposed on coplanar orbits, each debris-to-debris trajectory can be parameterized by $7$ variables, one being the final encounter time and the other $6$ identifying the trajectory arcs making up the transfer. So, to optimize the whole chaser trajectory, \textit{EOS} must deal with a $7N$-variable problem.
Because of the high problem dimensions (for typical values of $N$) the island-model parallelization of \textit{EOS} is decisive to attain good quality results is a limited amount of time.

\begin{table}[htbp]
\caption{Results of the application of \textit{EOS} to the $15$-target ADR mission by varying the number of islands.}
\begin{center}
\begin{tabular}{ccc|ccc}
\hline
\multicolumn{3}{c}{Algorithm parameters} & \multicolumn{3}{c}{Fitness stats, [\si{\kilo\meter/\second}]} \\ 
\hline
$N_i$ & $N_p$ & $N_G$ & $mean$  & $std$ & $best$ \\ 
\hline
$1$ & $512$ & $20000$ & $0.8400$  & $0.0755$ & $0.7310$ \\ 
$4$ & $256$ & $10000$ & $ 0.8165$  & $0.0842$ & $0.7173$ \\ 
$8$ & $128$ & $10000$ & $0.7577$  & $0.0444$ & $0.6779$ \\ 
$12$ & $85$ & $10000$ & $0.7575$  & $0.0408$ & $0.6900$ \\ 
$16$ & $64$ & $10000$ & $0.7475$  & $0.0343$ & $0.6957$ \\ 
$20$ & $51$ & $10000$ & $0.7369$  & $0.0458$ & $0.6668$ \\ 
$24$ & $43$ & $10000$ & $0.7231$  & $0.0228$ & $0.6935$ \\ 
$28$ & $37$ & $10000$ & $0.7389$  & $0.0437$ & $0.6816$ \\ 
$32$ & $32$ & $10000$ & $0.7381$  & $0.0318$ & $0.6768$ \\ 
\hline
\end{tabular}
\label{tab:adr}
\end{center}
\end{table}

\begin{figure*} [hbtp]
    \centering
    \begin{subfigure}[b]{0.45\textwidth}
        \centering
        \includegraphics[width=0.7\textwidth]{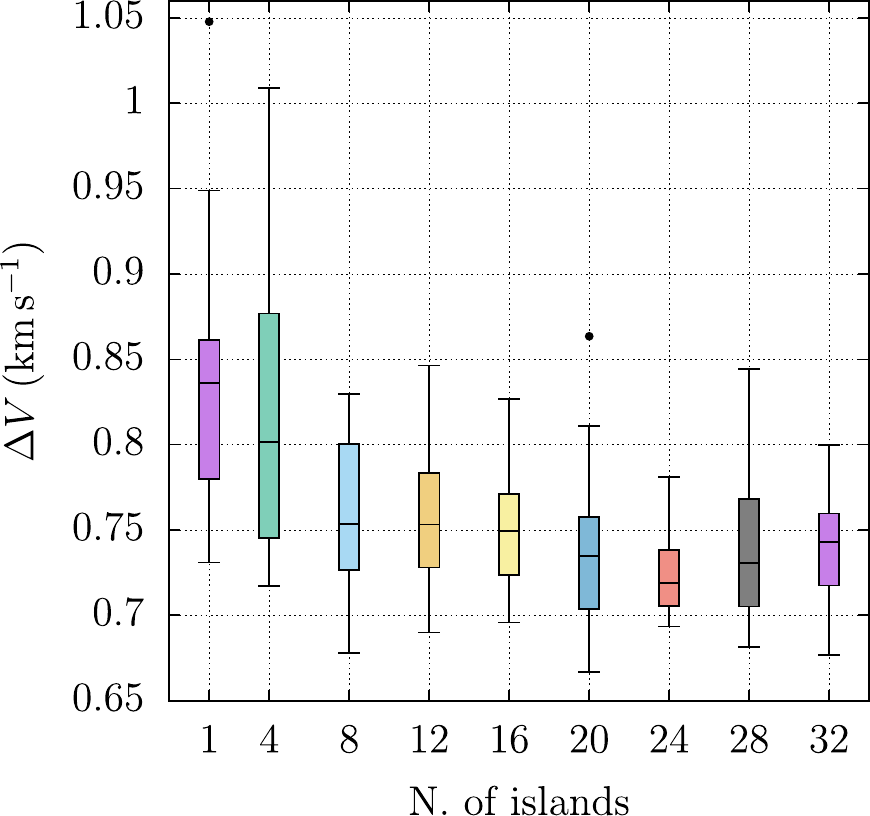}
        \caption{Fitness (total $\Delta V$) vs number of islands.}
        \label{fig:stats_fit}
    \end{subfigure}
    \begin{subfigure}[b]{0.45\textwidth}
        \centering
        \includegraphics[width=0.7\textwidth]{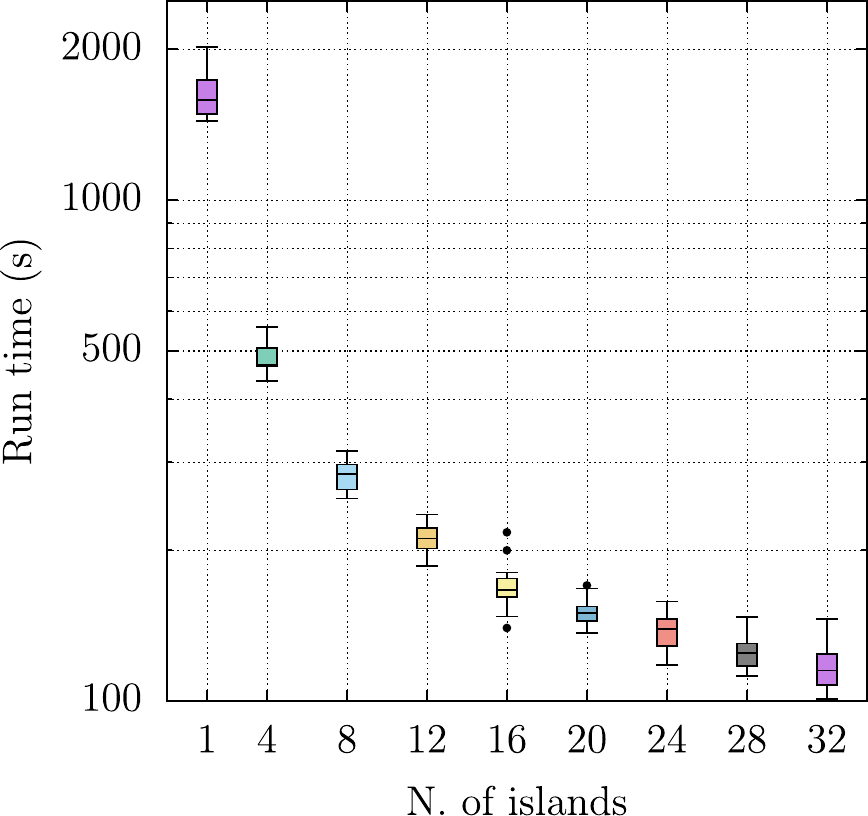}
        \caption{Run time (logarithmic scale) vs number of islands.}
        \label{fig:stats_time}
    \end{subfigure}
       \caption{Boxplots of run time and solution quality as a function of the number of islands $N_i$ for a $15$-target ADR mission. 
       Statistics are evaluated over $20$ independent runs, with a maximum FES $N_F = 10240000$.}
       \label{fig:stats_ADR}
\end{figure*}

Figure~\ref{fig:stats_ADR} shows a few results for a $15$-debris ADR mission ($105$ variables) as a function of the 
number of islands $N_i$, always considering approximately the same maximum number of FES.
The corresponding population size and number of generations for each run are reported in Table~\ref{tab:adr}. The parameters $\phi_{mig} = 0.5$, $N_{mig} = 100$ and $\rho_{mig} = 0.05$ were used in these tests.
For the sake of comparison, the results obtained through a single-island version of \textit{EOS}, with self-adaptive mutation strategy, are also reported.
All computations were performed on
KNL partition of CINECA's supercomputer Marconi, whose characteristics are reported in Sec.~\ref{sec:VEGA}.
%
%
The reduction of the overall run-time with the number of islands is apparent in Fig~\ref{fig:stats_time}, being a direct consequence of the parallel implementation of the algorithm. Moreover, as shown in Fig~\ref{fig:stats_fit}, the quality of the obtained solutions improves with the number of islands, up to $N_i=24$; from that point on, the population size of each island probably becomes too small for the problem at hand.


\section{Conclusion}
\label{sec:conclusion}

This paper presents the evolutionary optimization code \textit{EOS} developed at Sapienza University of Rome,
which represents a state-of-the-art DE-based algorithm.
After recalling the traditional DE algorithm, 
%
the main features added to \textit{EOS} to enhance the performance of DE were described in detail. These features include: the self-adaptation scheme for the DE control parameters, 
to avoid a manual tuning of such quantities and let them automatically evolve along the search; a partial restart (``epidemic'') mechanism, to enhance population diversity and evade the danger to get the search trapped in local optima; a pruning technique based on clustering, to focus the search on the most promising regions of the solution space; a $\varepsilon$-constrained method to make the algorithm capable of efficiently tackling constrained problems; a parallel synchronous island-model paradigm, to properly balance the ``explorative'' and ``exploitative'' tendencies of the algorithm and, at the same time, being able to distribute the computational load on a highly parallel environment, with a reduction of computational times and an increase in performance as immediate results.
\textit{EOS} demonstrated to be successful when applied to hard, real-world (unconstrained and constrained) problems of space trajectory design, characterized by many variables and the presence of several local minima, such as multiple gravity-assist trajectories, the ascent trajectory of a rocket, and multi-target rendezvous missions.

\bibliographystyle{LaTeX_Bibliography_Files/IEEEtran}
\bibliography{bibliography}

\end{document}